\documentclass[letterpaper, 10pt, conference]{ieeeconf}  
\IEEEoverridecommandlockouts                              
\usepackage[applemac]{inputenc}
\usepackage{graphics} 
\usepackage{epsfig} 
\usepackage{epstopdf}
\usepackage{algorithm}
\usepackage{algpseudocode}
\usepackage{todonotes}
\usepackage{amsmath} 
\usepackage{amssymb}  
\usepackage{epstopdf}
\usepackage{graphicx,xcolor}
\usepackage{caption}
\usepackage{subcaption}
\usepackage{courier}
\usepackage{mathrsfs}
\usepackage{blindtext}

\usepackage{pgfplots}
\pgfplotsset{compat=newest}
\pgfplotsset{plot coordinates/math parser=false}
\newlength\figureheight
\newlength\figurewidth

\newtheorem{thm}{\textbf{Theorem}}
\newtheorem{defi}{\textbf{Definition}}

\newtheorem{problem}{\textbf{Problem}}

\newtheorem{prop}{\textbf{Proposition}}

\title{\LARGE \bf
Safe Policy Synthesis in Multi-Agent POMDPs \\ via Discrete-Time  Barrier Functions}

\author{Mohamadreza Ahmadi, Andrew Singletary, Joel W. Burdick, and Aaron D. Ames
\thanks{M. Ahmadi, A. Singletary, J. W. Burdick, and A. D. Ames are with the California Institute of Technology, 1200 E. California Blvd., MC 104-44, Pasadena, CA 91125,  e-mail: (\{mrahmadi, asinglet, jwb, ames\}@caltech.edu). 
}
}

\begin{document}

\maketitle
\thispagestyle{empty}
\pagestyle{empty}

\begin{abstract}

A multi-agent partially observable Markov decision process (MPOMDP) is a modeling paradigm used for high-level planning of heterogeneous autonomous agents subject to uncertainty and partial observation. Despite their modeling efficiency, MPOMDPs have not received significant attention in safety-critical settings. In this paper, we use  barrier functions  to design policies for MPOMDPs that ensure safety. Notably, our method does not rely on  discretizations  of the belief space, or finite memory. To this end, we formulate  sufficient and necessary conditions for the safety of a given set based on discrete-time barrier functions (DTBFs) and we demonstrate that our formulation also allows for Boolean compositions of DTBFs for representing more complicated safe sets. We show that the proposed method can be implemented online by a sequence of one-step greedy algorithms as a stand-alone safe controller or as a safety-filter given a nominal planning policy. We illustrate the efficiency of the proposed methodology based on DTBFs using a high-fidelity simulation of  heterogeneous~robots.

\end{abstract}

\section{INTRODUCTION}

\begin{figure}
    \centering
    \includegraphics[width=5cm]{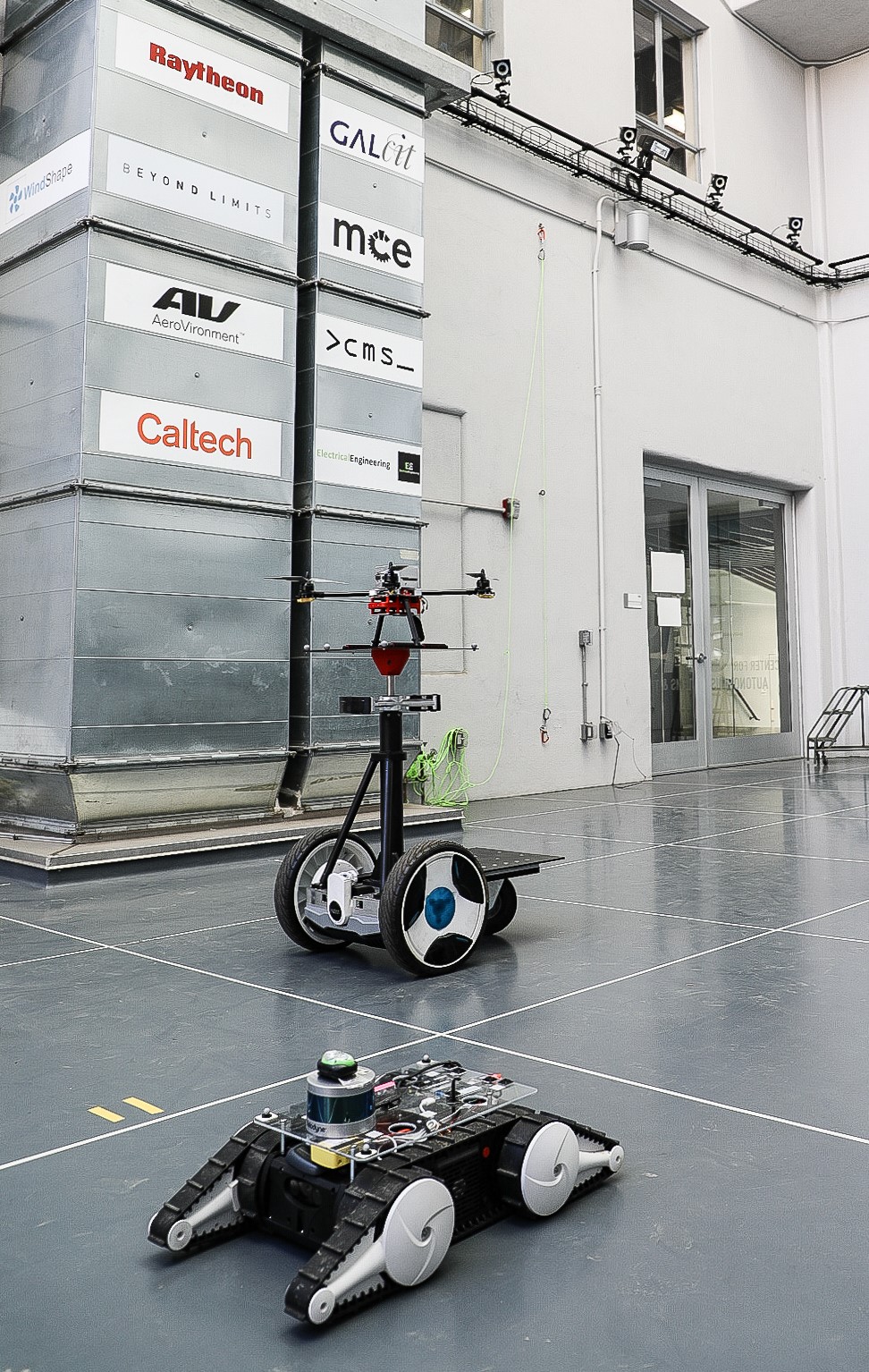}
    \caption{{A team of heterogeneous robots consisting of a quadrotor, a Segway, and a Flipper.}}
    \label{fig:proposal}
\end{figure}

Complex mission planning
of multiple heterogeneous robots, e.g. flying and ground robots (see Figure~\ref{fig:proposal}), presents an inherent tension between the need for
greater autonomy and the absolute necessity of strong safety~\cite{ahmadi2017safety}  and performance guarantees~\cite{ahmadi2014guaranteed}. Safety is crucial for the duration of a safety-critical mission, for example, those involving human-robot interactions~\cite{vasic2013safety}. The planning problem becomes even more involved in the presence of partial or uncertain information
about the world, as well as stochastic actions and noisy sensors~\cite{pynadath2002communicative,seuken2008formal}.

Multi-agent partially observable Markov decision processes~\cite{messias2011efficient,amato2015scalable} provide a sequential  decision-making formalism for high-level planning of multiple autonomous agents under partial observation and uncertainty. In MPOMDPs, the agents share their local observations and make decisions based on a global information state (the joint belief). Despite this unique modeling paradigm, the computational complexity of MPOMDPs is PSPACE-complete~\cite{bernstein2002complexity,hansen2004dynamic}. Therefore, several promising approximate methods for solving MPOMDPs have been proposed in the literature, e.g. sampling-based methods~\cite{amato2015scalable} and point-based methods~\cite{shani2013survey}. However, it is difficult to provide safety assurances when one employs approximate methods for solving MPOMDPs, as such  methods either use discretization techniques~\cite{haesaert2018temporal} or finite-state controllers~\cite{sharan2014finite}. 


Safety verification can be encoded as checking whether the solutions of a system remain inside a pre-specified safe set or alternatively avoid a pre-defined unsafe set. Then, a natural method for checking safety is to compute the reachable set of a system subject to disturbances and controls~\cite{mitchell2005time,abate2008probabilistic,althoff2008reachability}. However, for complex and high-dimensional systems such methods are either intractable, or overly conservative. Alternative approaches to reachability date back to the pioneering works of Nagumo~\cite{nagumo1942lage} to study the set invariance of ordinary differential equations (ODEs). Nagumo's works were extended to ODEs with inputs by Aubin~\textit{et. al.}~\cite{aubin2011viability} in the context of viability theory. The interest in hybrid systems in the 2000's led to the introduction of barrier certificates for safety verification~\cite{prajna2007framework}. However, the construction of these barrier certificates require solving a set of polynomial optimization problems that become intractable for high-dimensional systems (despite some promising recent  directions~\cite{ahmadi2014dsos}). The recently proposed notion of barrier functions~\cite{ames2017control} circumvent the computational bottleneck of barrier certificates inasmuch as the closed-form expression for a barrier function can be derived from the definition of the safe set. By taking advantage of this property, barrier functions have been used for designing safe controllers (in the absence of a nominal controller) and safety filters (in the presence of a nominal controller) for dynamical systems, such as biped robots~\cite{nguyen20163d} and trucks~\cite{chen2017enhancing}, with guaranteed performance and robustness~\cite{xu2015robustness,kolathaya2019input}.

In this paper, we extend the application of barrier functions from low-level safety constraints of dynamical systems to high-level safety objectives of MPOMDPs. Our results are based on the observation that  the joint belief evolution of an MPOMDP is described by a discrete-time system~\cite{ahmadi2018privacy,ahmadi2018verification,ahmadi2018barrier}. We begin by formulating a,  both necessary and sufficient, theorem for safety verification of a given set for discrete-time systems based on \emph{discrete-time barrier functions} (DTBFs) and we demonstrate that our formulation  allows for more complicated safe belief sets described by Boolean compositions of DTBFs. Then, we apply these DTBFs to study the safety of a given set of safe beliefs. We propose online methods based on one-step greedy algorithms to either synthesize a safe policy for an MPOMDP or synthesize a safety filter for an MPOMDP given a nominal planning policy. We illustrate the efficacy of the proposed approach by applying it to an exploration scenario of a team of heterogeneous robots in a high-fidelity simulation environment. 

The rest of the paper is organized as follows. In the next section, we briefly review MPOMDPs and related notions. In Section~\ref{sec:BF}, we formulate a barrier function theorem for discrete-time systems. In Section~\ref{sec:safempomdp}, we use the tools developed in Section~\ref{sec:BF} to ensure safe planning in MPOMDPs.  Section~\ref{sec:example} elucidate our results by a high-fidelity multi-robot exploration simulation. Finally, in Section~\ref{sec:conclude}, we conclude the paper and give directions for future reserch.



\vspace{0.3cm}
\textbf{Notation:}
$\mathbb{R}^n$ denotes the $n$-dimensional Euclidean space. $\mathbb{R}_{\ge 0}$ denotes the set $[0,\infty)$. $\mathbb{N}_{\ge0}$ denotes the set of non-negative integers. For a finite-set $A$,  $|A|$ denotes the number of elements in $A$. A continuous function $f:[0,a)\to \mathbb{R}_{\ge 0}$ is a class $\mathcal{K}$ function if $f(0)=0$ and it is strictly increasing. Similarly,  a continuous function $g:[0,a)\times \mathbb{R}_{\ge 0} \to \mathbb{R}_{\ge 0}$ is a class $\mathcal{KL}$ function if $g(r,\cdot) \in \mathcal{K}$ and if  $g(\cdot,s)$ is decreasing with respect to $s$ and  $\lim_{s \to \infty} g(\cdot,s) \to 0$. For two functions $f:\mathcal{G}\to \mathcal{F}$ and $g:\mathcal{X}\to \mathcal{G}$, $f \circ g:\mathcal{X} \to \mathcal{F}$ denotes the composition of $f$ and $g$ and $\mathrm{Id}:\mathcal{F} \to \mathcal{F}$ denotes the identity function satisfying $\mathrm{Id}\circ f = f$ for all functions~$f:\mathcal{X} \to \mathcal{F}$.

\section{Multi-Agent POMDPs}

An MPOMDP~\cite{messias2011efficient,amato2015scalable} provides a sequential  decision-making formalism for high-level planning of multiple autonomous agents under partial observation and uncertainty. At every time step, the agents take actions and receive observations. These observations are shared via (noise and delay free) communication and the agents decide in a centralized framework.
\vspace{0.3cm}
\begin{defi}
\textit{
An MPOMDP is a tuple $$\left(I, Q,p^0, \{A_i\}_{i\in I},T,R,\{Z_i\}_{i \in I},O\right),$$ wherein
	\begin{itemize}
        \item $I$ denotes a index set of agents;
		\item $Q$ is a finite set of states with indices $\{1,2,\ldots,n\}$ (which can be described as the product space of the states of all agents);
		\item $p^0:Q\rightarrow[0,1]$ defines the distribution of the initial states, i.e., $p^0(q)$ denotes the probability of starting at $q\in Q$;
		\item $A_i$ is a finite set of actions for agent $i$ and $A=\times_{i\in I}A_i$ is the set of joint actions;
		\item $T:Q\times A\times Q\rightarrow [0,1]$ is the transition probability, where $
		T(q,a,q'):=P(q^t=q'|q^{t-1}=q,a^{t-1}=a),~\\
		\forall t\in\mathbb{Z}_{\ge 1}, q,q'\in Q, a\in A, $
i.e., the probability of moving to state $q'$ from $q$ when the joint actions $a$ are taken;
\item $R:Q \times A\rightarrow \mathbb{R}$	is the immediate reward function for taking the joint action $a$ at state $q$;
    		\item $Z_i$ is the set of all possible observations for agent $i$ and $Z=\times_{i \in I} Z_i$, representing outputs of  discrete sensors. Often, $z\in Z$ are incomplete projections of the world states $q$, contaminated by sensor noise;
		\item $O:Q\times A \times Z\rightarrow [0,1]$ is the observation probability (sensor model), where $
		O(q',a,z):=P(z^t=z|q^{t}=q',a^{t-1}=a),~
	    \forall t\in\mathbb{Z}_{\ge 1}, q\in Q, a\in A, z\in Z, $
i.e., the probability of seeing joint observations $z$ given joint actions $a$ were taken and resulting in state $q'$.
	\end{itemize}
	}
\end{defi}
\vspace{0.3cm}

 	Since the states are not directly accessible in an MPOMDP, decision making requires the history of joint actions and joint observations. Therefore, we must define the notion of a joint \emph{belief} or the posterior as sufficient statistics for the history~\cite{astrom1965optimal}. Given an MPOMDP, the joint belief at $t=0$ is defined as $b^0(q)=p^0(q)$ and $b^t(q)$ denotes the probability of the system being in state $q$ at time $t$. At time $t+1$, when joint action $a\in A$ is taken and joint observation $z \in Z$ is observed, the belief is updated via a Bayesian filter as

\begin{align} 
& b^t(q') \label{equation:belief update}
= P(q'|z^t,a^{t-1},b^{t-1})   \nonumber\\
&= \frac{P(z^t|q',a^{t-1},b^{t-1})P(q'|a^{t-1},b^{t-1})}{P(z^t|a^{t-1},b^{t-1})}\nonumber \\
&= \frac{P(z^t|q',a^{t-1},b^{t-1})}{P(z^t|a^{t-1},b^{t-1})}  \nonumber\\
&\quad   \times \sum_{q\in Q}P(q'|a^{t-1},b^{t-1},q)P(q|a^{t-1},b^{t-1}) \nonumber \\
&= \frac{O(q',a^{t-1},z^{t})\sum_{q\in Q}T(q,a^{t-1},q')b^{t-1}(q)}{\sum_{q'\in Q}O(q',a^{t-1},z^{t})\sum_{q\in Q}T(q,a^{t-1},q')b^{t-1}(q)}
\end{align}

where the beliefs belong to the belief unit simplex
$$
\mathcal{B} = \left\{ b \in [0,1]^{|Q|} \mid \sum_{q\in Q} b^t(q)=1,~\forall t  \right\}.
$$
A policy in an MPOMDP setting is then a mapping $\pi:\mathcal{B} \to A$, i.e., a mapping from the continuous joint beliefs space into the discrete and finite joint action space. The special case of $I$ being a singleton (only one agent) is known as a partially observable Markov decision process (POMDP)~\cite{smallwood1973optimal}.

The execution of an MPOMDP is  carried out in the following steps~\cite{oliehoek2012tree}. At every time step $t$, each agent~$i$ observes $z_i^t$ and communicates its own observation $z_i^t$ to all other agents. The agent then in return receives observations of others $z\setminus\{z_i\}$ and  uses the joint observations $z^t$ and the previous joint action $a^{t-1}$ to update the new joint belief $b^t$ from~\eqref{equation:belief update}. Finally, the agent looks up the joint action from the joint policy $\pi(b^t) = a^t$ and executes its individual action~$a_i^t$.

Noting that the joint belief evolution of an MPOMDP~(1) is described by a discrete-time system~\cite{ahmadi2018privacy,ahmadi2018verification,ahmadi2018barrier}, in the next section, we propose conditions based on DTBFs for safety analysis of discrete-time systems.

\section{barrier functions for\\ Discrete-Time Systems}\label{sec:BF}

While there is a long history of studying the set invariance properties of dynamical systems \cite{nagumo1942lage}, recently these concepts were extended to include conditions over a set.  This was done through the concepts of barrier functions~\cite{Ames2017}. In the same vein, in~\cite{agrawal2017discrete}, the barrier function method was extended to discrete-time dynamical systems. Unfortunately, with the latter formulation of the (reciprocal) barrier functions, we can not study the solutions of the discrete-time system outside of the invariant set, i.e., if the solution is on the boundary of the set or when it leaves the set. To overcome this difficulty, we next extend the notion of (zeroing) barrier functions~\cite{Ames2017} to discrete-time systems.


\subsection{Discrete-Time Barrier Functions}

We consider the following discrete-time system
\begin{equation}\label{eq:discdyn}
x^{t+1} = f(x^{t}),~~t \in \mathbb{N}_{\ge 0},
\end{equation}
with $f:\mathcal{X}  \to \mathcal{X} \subset \mathbb{R}^n$ and a safe set defined as
\begin{eqnarray}\label{eq:safeset}
\mathcal{S} :=\{ x \in \mathcal{D} \mid h(x) \ge 0 \}, \\
\mathrm{Int}(\mathcal{S}) :=\{ x \in \mathcal{D} \mid h(x) > 0 \}, \\
\partial \mathcal{S} :=\{ x \in \mathcal{D} \mid h(x) = 0 \}.
\end{eqnarray}
\vspace{.3cm}
We then have the following definition of a DTBF.
\begin{defi}[Discrete-Time Barrier Function]
\textit{
For~the discrete-time system~\eqref{eq:discdyn}, the continuous function $h : \mathbb{R}^n \to \mathbb{R}$ is a discrete-time barrier
function for the set $\mathcal{S}$ as defined in (3)-(5), if there exists $\alpha \in \mathcal{K}$ satisfying $\alpha(r) < r$ for all $r>0$ and a set $\mathcal{D}$ with $\mathcal{S} \subseteq \mathcal{D} \subset \mathbb{R}^n$ such that
\begin{equation}\label{eq:BFinequality}
    h(x^{t+1})-h(x^{t}) \ge - \alpha(h(x^{t})),\quad \forall x \in \mathcal{D}.
    \end{equation}
    }
    \end{defi}
\vspace{.3cm}

In fact, the discrete-time barrier function would more correctly be called a discrete-time zeroing barrier function per the literature \cite{ames2017control}, but we drop the ``zeroing'' as it is the only form of barrier function that will be considered throughout the rest of this paper.

We can show that the existence of a DTBF is both necessary and sufficient for invariance. 

\vspace{.3cm}
\begin{thm}\label{thm:BFdiscrete}
\textit{
Consider the discrete-time system~\eqref{eq:discdyn}. Let $\mathcal{S} \subseteq \mathcal{D} \subset \mathbb{R}^n$ with $\mathcal{S}$ as described in (3)-(5). Then, $\mathcal{S}$ is invariant if and only if there exists a DTBF as defined in Definition 2.
}
\end{thm}
\vspace{.3cm}
\begin{proof}
We begin by proving the sufficiency. If~\eqref{eq:BFinequality} holds, we have $
h(x^{t}) \ge (\mathrm{Id}-\alpha) \circ h(x^{t-1})$.
Furthermore, since $\alpha(r)\le r$, $(\mathrm{Id}-\alpha) \circ (r) < r$ and $(\mathrm{Id}-\alpha) \in \mathcal{K}$~\cite{jiang2002converse}. For $t=0$, we have $$h(x^1)\ge (\mathrm{Id}-\alpha)\circ h(x^0).$$
Similarly, for $t=1$, we have $$h(x^2)\ge (\mathrm{Id}-\alpha)\circ h(x^1).$$

From the inequality obtained at $t=0$, we obtain $h(x^2)\ge  (\mathrm{Id}-\alpha) \cdot h(x^1) \ge (\mathrm{Id}-\alpha)\circ (\mathrm{Id}-\alpha) \circ h(x^0)$. Then, by induction, we conclude \begin{equation}\label{eq:BFrate}h(x^{t}) \ge (\mathrm{Id}-\alpha)^t \circ h(x^0),\quad t\in \mathbb{N},\end{equation} where $(\mathrm{Id}-\alpha)^t$ denotes composition $t$ times.

At this point, we check invariance of $\mathcal{S}$ and asymptotic convergence (followed by invariance) of solutions to $\mathcal{S}$  for the two cases of $x^0 \in \mathcal{S}$ and $x^0 \in \mathcal{D}\setminus \mathcal{S}$, respectively.

For any $x^0 \in \mathcal{S}$, since $h(x^0)\ge 0$ by definition of $\mathcal{S}$ and $(\mathrm{Id}-\alpha) \in \mathcal{K}$, we can deduce from~\eqref{eq:BFrate} that $h(x^{t})\ge 0$ for all $t \in \mathbb{N}$, implying that $\mathcal{S}$ is invariant. This is simply because if $(\mathrm{Id}-\alpha) \circ (r) < r$, then there exist a constant $\gamma \in (0,1)$ such that $(\mathrm{Id}-\alpha) \circ (r) \le  \gamma r$ and hence $(\mathrm{Id}-\alpha)^t \circ (r) \le \gamma^t r$.

For any $x^0 \in \mathcal{D}\setminus \mathcal{S}$, inequality~\eqref{eq:BFrate} implies that as $t \to \infty$, we have $h(x^{t})\ge 0$. That is, all solutions of system~\eqref{eq:discdyn} starting at $x^0 \in \mathcal{D}\setminus \mathcal{S}$, asymptotically converge to  $\mathcal{S}$.

We next prove the converse direction. We set $\mathcal{S}=\mathcal{D}$ in Theorem 1. If $\mathcal{S}$ is forward invariant, we have $x^{t-1} \in \mathcal{S}$ and $x_{t} \in \mathcal{S}$ for all $t \in \mathbb{N}$. From the definition of the set $\mathcal{S}$, this implies that if $h(x^{t-1})\ge 0$ then $h(x^{t})\ge 0$ for all $t \in \mathbb{N}$. Furthermore, we claim if $\mathcal{S}$ is forward invariant, then $h(x^{t})-h(x^{t-1})\ge 0$. Because if $h(x^{t})-h(x^{t-1})\le 0$ or alternatively $h(x^{t})\le h(x^{t-1})$, for all $x^{t-1} \in \partial \mathcal{S}$, we have $h(x^{t})\le 0$. That is, $x^{t} \notin \mathcal{S}$ which is a contradiction. Hence, we have $h(x^{t})-h(x^{t-1})\ge 0$  for all $t \in \mathbb{N}$.

For any $r\ge 0$, the set $\{ x \in \mathbb{R}^n \mid 0 \le h(x) \le r\}$ is a compact subset of $\mathcal{S}$. Define a function $\alpha : [0,\infty) \to \mathbb{R}$ by
$$
\alpha(r) = - \inf_{\{ x' \mid 0 \le h(x') \le r\}}\inf_{\{ x \mid 0 \le h(x) \le r\}} \left( h(x')-h(x) \right).
$$
Using the compactness property stated above and the fact that the difference of two continuous functions is continuous, $\alpha$ is a well-defined, non-decreasing function on $\mathbb{R}_{\ge 0}$ satisfying
$$
 h(x^{t})-h(x^{t-1}) \ge - \alpha \circ h(x^{t-1}), \quad \forall x^{t-1} \in \mathcal{S}.
$$
Moreover, if $\mathcal{S}$ is forward invariant,  $h(x^{t})\ge 0$ for all $t \in \mathbb{N}$. That is,  $h(x^{t})\ge (\mathrm{Id}  - \alpha) \circ h(x^{t-1})$. Since $h(x^{t-1})\ge 0$, $(\mathrm{Id}-\alpha)\cdot (r) > 0 $, which implies $\alpha(r) < r$. This completes the proof.
\end{proof}
\vspace{0.3cm}
Note that a simple example of the function $\alpha$ in inequality~\eqref{eq:BFinequality} is when $\alpha$ is a constant $\alpha_0\in (0,1)$. In this case, from the proof of Theorem 1, we infer that
$$h(x^{t}) \ge (1-\alpha_0)^t h(x^0),\quad t\in \mathbb{N}.$$
Indeed, we can control the rate of convergence of the DTBF by changing the value of $\alpha_0$.

As a technical remark, we point out that, unlike proving the converse BF theorem for continuous-time systems~\cite{ames2017control}, we did not invoke Nagumo's theorem on the boundary of the set $\mathcal{S}$. This is simply because such condition does not imply invariance for discrete-time systems~\cite[Section 3.2]{blanchini1999set}.

\subsection{Boolean Composition of Discrete-Time Barrier Functions}

It is often desirable to consider sets defined by Boolean composition of multiple barrier functions. In this regard, in~\cite{glotfelter2017nonsmooth}, the authors proposed non-smooth barrier functions as a means to analyze composition of barrier functions by Boolean logic, i.e., $\lor$ (disjunction), $\land$ (conjunction), and $\neg$ (negation). Similarly, in this study,  we use $\max$ to represent $\lor$ and $\min$ to show $\land$. In other words, if $x \in \{x \in \mathbb{R}^n \mid \max_{i=1,\ldots,k} h_i(x) \ge 0\}$, then there exists at least one $i_* \in \{1,\ldots,k\}$ such that $h_{i_*}(x) \ge 0$ and if $x \in \{x \in \mathbb{R}^n\mid \min_{i=1,\ldots,k} h_i(x) \ge 0\}$, then for all $i \in \{1,\ldots,k\}$ we have $h_{i_*}(x) \ge 0$. The negation operator is trivial and can be shown by checking if $-h$ satisfies the invariance property.

In the following, we propose conditions for checking Boolean compositions of barrier functions. Fortunately, since we are concerned with discrete time systems, this does not require non-smooth analysis.

In the context of DTBFs, we have the following result. 
\vspace{.3cm}
\begin{prop}
\textit{
Let $\mathcal{S}_i = \{ x \in \mathbb{R}^n \mid h_i(x)\ge 0\}$, $i=1,\ldots,k$ denote a family of safe sets with the boundaries and interior defined analogous to $\mathcal{S}$ in~\eqref{eq:safeset}. Consider the discrete-time system~\eqref{eq:discdyn}. If there exist a $\alpha \in \mathcal{K}$ satisfying $\alpha(r) < r$ for $\forall r>0$ such that
\begin{equation}\label{eq:disjBF1}
\min_{i=1,\ldots,k} h_i(x^{t+1}) - \min_{i=1,\ldots,k} h_i(x^{t}) \ge - \alpha\left(\min_{i=1,\ldots,k}  h_i(x^{t})\right)
\end{equation}
then the set $\{ x \in \mathbb{R}^n \mid  \land_{i =1,\ldots,k} h_i(x) \ge 0\}$ is forward invariant. Similarly, if there exist a $\alpha \in \mathcal{K}$ satisfying $\alpha(r) < r$ for  all $ r>0$ such that
\begin{equation}\label{eq:disjBF2}
\max_{i=1,\ldots,k} h_i(x^{t+1}) - \max_{i=1,\ldots,k} h_i(x^{t}) \ge - \alpha\left(\max_{i=1,\ldots,k}  h_i(x^{t})\right)
\end{equation}
then the set $\{ x  \in \mathbb{R}^n \mid  \lor_{i =1,\ldots,k} h_i(x) \ge 0\}$ is forward invariant.
}
\end{prop}
\vspace{.3cm}
\begin{proof}
We prove the case for conjunction and the proof for disjunction is similar. If~\eqref{eq:disjBF1} holds from the proof of Theorem 1, we can infer that 
$$
\min_{i=1,\ldots,k} h_i(x^{t}) \ge (\mathrm{Id}-\alpha)^t\circ \left(\min_{i=1,\ldots,k} h_i(x^0)\right).
$$
That is, if $x^0 \in \{ x \in \mathbb{R}^n \mid \min_{i=1,\ldots,k} h_i(x) \ge 0\}$, then $\min_{i=1,\ldots,k} h_i(x^{t}) \ge 0$ for all $t\in \mathbb{N}_{\ge 0}$, which in turn implies that  $h_{i}(x) \ge 0$ for all $i \in \{1,\ldots,k\}$.
\end{proof}
\vspace{.3cm}
The next section shows how the results in this section can be used to provide safety assurances for MPOMDPs.

\section{Safety-Critical Control of MPOMDPs}\label{sec:safempomdp}

Since the states are not directly observable in MPOMDPs, we are interested in guaranteeing safety in a probabilistic setting in the joint belief space. To this end, we define the set of safe joint beliefs as
\begin{eqnarray}\label{eq:safesetbelief}
\mathcal{B}_s :=\{ b  \in \mathcal{B} \mid h(b) \ge 0 \},\\
\mathrm{Int}({B}_s) :=\{ b  \in \mathcal{B} \mid h(b) > 0 \},\\
\partial \mathcal{B}_s :=\{ b \in \mathcal{B} \mid h(b) = 0 \},
\end{eqnarray}
where $h:\mathcal{B} \to \mathbb{R}$ is a given function. We denote by $\pi_n : \mathcal{B} \to \mathcal{A}$ a nominal joint policy mapping each joint belief into a joint action. We use subscript $n$ to denote variables corresponding to the nominal policy designed offline.

We are interested in solving the following problems for MPOMDPs.
\vspace{.3cm}
\begin{problem}
\textit{
Given an MPOMDP as defined in Definition~1, a corresponding belief update~\eqref{equation:belief update}, and a safe joint belief set $\mathcal{B}_s$, design a sequence of actions $a^t,~t \in \mathbb{N}_{\ge 0}$ such that $b^t \in \mathcal{B}_s,~\forall t\in\mathbb{N}$ and the instantaneous rewards  $r^t = \sum_{q^t \in Q} b(q^t) R(q^t,a^t)$ are maximized for all~$t \in \mathbb{N}_{\ge 0}$.
}
\end{problem}
\vspace{.3cm}
\begin{problem}
\textit{
Given an MPOMDP as defined in Definition~1, a corresponding belief update equation~\eqref{equation:belief update},  a safe joint belief set $\mathcal{B}_s$, and a nominal planning policy $\pi_n$, determine a sequence of actions $a^t,~t \in \mathbb{N}_{\ge 0}$ such that $b^t \in \mathcal{B}_s,~\forall t\in\mathbb{N}_{\ge 0}$ and the quantity   $\|r^t - r_n^t\|^2$ is minimized for all $t \in \mathbb{N}_{\ge 0}$, where $r_n^t$ denotes the nominal immediate reward at time step $t$.
}
\end{problem}
\vspace{.3cm}
As can be inferred from Problems 1 and 2, we seek to ensure safety in addition to motion planning at every time step. Such problems are prevalent in multi-agent robot applications, where safety is of significant importance, e.g., robots in performing tasks in the presence of  human coworkers~\cite{murashov2016working}.

\subsection{Barrier Functions for MPOMDPs}

Next, we use the result in Theorem~1 to  ensure safety of a team of heterogeneous autonomous agents described by an MPOMDP. To this end, we solve the following discrete optimization problem at each time step $t$:

\begin{align}
a^* =  \arg \max_{a \in A}~~ & \left(\sum_{q' \in Q} b(q') R(q',a)\right) \nonumber \\ {s.t.}~~ &h(b(q'))-h(b(q)) \ge - \alpha(b(q)).
\end{align}

\begin{algorithm}[t]
    \begin{algorithmic}
        \Require System information $I$, $Q$, $A$, $T$, $R$, $Z$, $O$, safe belief set $\mathcal{B}_s$, current observation $z^t$, the past belief $b^{t-1}$
        \Statex
        \State  $i=1$ 
         \For{$i =1,2,\ldots,|A|$}
\State $ b^t(q') = \frac{O(z^t\mid q',a(i))\sum_{q \in Q}T(q'\mid q,a(i))b^{t-1}(q)}{\sum_{q'\in Q}O(z^t\mid q',a(i))\sum_{q \in Q}T(q'\mid q,a(i))b^{t-1}(q)}$ \\
  \If{$    h(b^{t})-h(b^{t-1}) \ge - \alpha(h(b^{t-1}))$} 
 \State $ r(i) = \left(\sum_{q' \in Q} b(q') R(q',a(i))\right)$ 
 \EndIf
  \EndFor
\State $i_* = \arg \max_{i=1,2,\ldots,|A|} r(i)$ 
\Statex
 \Return $a^* = a({i_*})$.
\Statex
    \end{algorithmic}
    \caption{The one-step greedy algorithm for finding the safe action at time $t$.}\label{alg:daclyf}
\end{algorithm}

Algorithm 1 summarizes the steps involved in finding the safe action based on barrier functions at each time step. At every time step, the algorithm picks a joint action $a(i)$  from $|A|$ combinations of actions (recall that $\times_{i \in I}A_i =A$). For each joint action $a(i)$, it computes the next joint belief and checks whether if the next joint belief satisfies the safety constraint. If the safety constraint is satisfied, it computes the corresponding reward function $r(i)$ for the joint action $a(i)$. After checking all actions, the algorithm returns the joint action maximizing the reward function.

Algorithm~1 designs a safe and mypoic optimal action at each time step based on the current observation and the belief state at the step before. Therefore, it does not require a full memory of past actions and observations. This synthesis algorithm for POMDPs parallels those using control barrier functions for dynamical systems wherein safety for all time and optimality at each time instance is required~\cite{ames2017control}.

Note that, if the safety requirement is defined by Boolean logic and we need to check either inequality~\eqref{eq:disjBF1} or~\eqref{eq:disjBF2}, we can just replace the inequality in the ``if'' statement in Algorithm~1 with either inequality~\eqref{eq:disjBF1} or~\eqref{eq:disjBF2}.


Furthermore, we remark that stability is not an issue in MPOMDP problems, since the beliefs evolve in the probabilistic belief simplex. However, we can encode instability in an MPOMDP problem as a set of bad states, that is, $\mathcal{B} \setminus \mathcal{B}_s$.

%

\begin{algorithm}[t]
    \begin{algorithmic}
        \Require System information $I$, $Q$, $A$, $T$, $R$, $Z$, $O$, safe belief set $\mathcal{B}_s$, current observation $z^t$, the past belief $b^{t-1}$
        \Statex
        \State  $i=1$ 
         \For{$i =1,2,\ldots,|A|$}
\State $ b^t(q') = \frac{O(z^t\mid q',a(i))\sum_{q \in Q}T(q'\mid q,a(i))b^{t-1}(q)}{\sum_{q'\in Q}O(z^t\mid q',a(i))\sum_{q \in Q}T(q'\mid q,a(i))b^{t-1}(q)}$ \\
  \If{$h_k(b^t_k(q'))-h_k(b^{t-1}_k(q)) \ge -\alpha_k(h_k(b^{t-1}_k(q))) $ for all $k \in I$} 
 \State $ r(i) = \left(\sum_{q' \in Q} b(q') R(q',a(i))\right)$ 
 \EndIf
  \EndFor
\State $i_* = \arg \max_{i=1,2,\ldots,|A|} r(i)$ 
\Statex
 \Return $a^* = a({i_*})$.
\Statex
    \end{algorithmic}
    \caption{The one-step greedy algorithm for finding the safe action at time $t$ when agents have different safety constraints.}\label{alg:daclyf}
\end{algorithm}

Each autonomous agent  might have a different safety requirement, characterized by sets $\mathcal{B}_i,~i \in I$, i.e., $\mathcal{B}_i$ is the safe set for agent $i$. Then, we just need to check the safety of each agent separately. We denote by $b_i$, $i \in I$, the subset of joint beliefs concerning  agent $i$, e.g. beliefs showing the location of the agent. Algorithm~2 demonstrates how we can  check the safety requirement of each agent separately at every time~step.

%

In many real world multi-robot navigation scenarios, an offline policy for path planning exists (e.g. based on point-based methods~\cite{shani2013survey}). However, such policy may not guarantee safety. We can use the barrier functions to design an online method for ensuring safety while remaining as much faithful as possible to the offline policy (see~\cite{borrmann2015control,gurriet2018online} for analogous formulations for systems described by nonlinear differential equations). 

\begin{algorithm}[t]
    \begin{algorithmic}
        \Require System information $I$, $Q$, $A$, $T$, $R$, $Z$, $O$, nominal policy $\pi_n$, safe belief set $\mathcal{B}_s$, current observation $z^t$, the past belief $b^{t-1}$
        \Statex
\State        $ b^t(q') = \frac{O(z^t\mid q',a_n^t)\sum_{q \in Q}T(q'\mid q,a_n^t)b^{t-1}(q)}{\sum_{q'\in Q}O(z^t\mid q',a_n^t)\sum_{q \in Q}T(q'\mid q,a_n^t)b^{t-1}(q)}$
\If{$    h(b^{t})-h(b^{t-1}) < - \alpha(h(b^{t-1}))$}
        \State  $i=1$ 
         \For{$i =1,2,\ldots,|A|$}
\State $ b^t(q') = \frac{O(z^t\mid q',a(i))\sum_{q \in Q}T(q'\mid q,a(i))b^{t-1}(q)}{\sum_{q'\in Q}O(z^t\mid q',a(i))\sum_{q \in Q}T(q'\mid q,a(i))b^{t-1}(q)}$ \\
  \If{$h(b^{t})-h(b^{t-1}) \ge - \alpha(h(b^{t-1}))$} 
 \State $ r(i) = \left(\sum_{q' \in Q} b(q') R(q',a(i))\right)$ 
 \EndIf
  \EndFor
\State $i_* =  \arg \min_{i=1,2,\ldots,|A|} \|r(i)-r_n^t\|^2$ 
\Statex
 \Return $a^* = a({i_*})$
\EndIf
\Statex
 \Return $a^* = a_n^t$.
\Statex
    \end{algorithmic}
    \caption{The one-step greedy algorithm for filtering the nominal policy with a safe action at every time-step $t$.}\label{alg:daclyf}
\end{algorithm}

 Algorithm 3 illustrates how barrier functions can filter the agent actions to ensure safety. At every time step $t$, the algorithm first computes the next joint belief $b^t$ given the nominal action $a_n$ designed based on the nominal policy $\pi_n$. It then checks whether that action leads to a safe joint belief update (this is allowed since the existence of a DTBF $h$ satisfying~\eqref{eq:BFinequality} is both necessary and sufficient for safety). If yes, the algorithm returns $a_n$ for implementation. If no, the algorithm finds a safe joint action that minimally changes the immediate reward from the nominal immediate reward $r_n^t$ in a least squares sense.


\section{Case Study: Multi-Robot Exploration} \label{sec:example}

To demonstrate our method, we consider a system of three heterogeneous robots exploring an unknown environment. The mission objective is to retrieve a sample located somewhere in the robots' vicinity. Each robot has different and limited capabilities to explore and observe the environment, so coordination and communication between the robots is required in order to complete the mission. 

The robot team consists of a drone and two ground vehicles. The drone can rapidly explore the environment from above, but is unable to explore any covered or underground regions. The ground vehicles include a Rover Robotics Flipper, and a modified Segway. The Flipper is a small, tracked vehicle capable of traversing in tight spaces and rough terrain, while the Segway a is larger, wheeled robot without external sensing capabilities, whose purpose is to retrieve the sample.

\begin{figure}
    \centering
    \includegraphics[width=8cm]{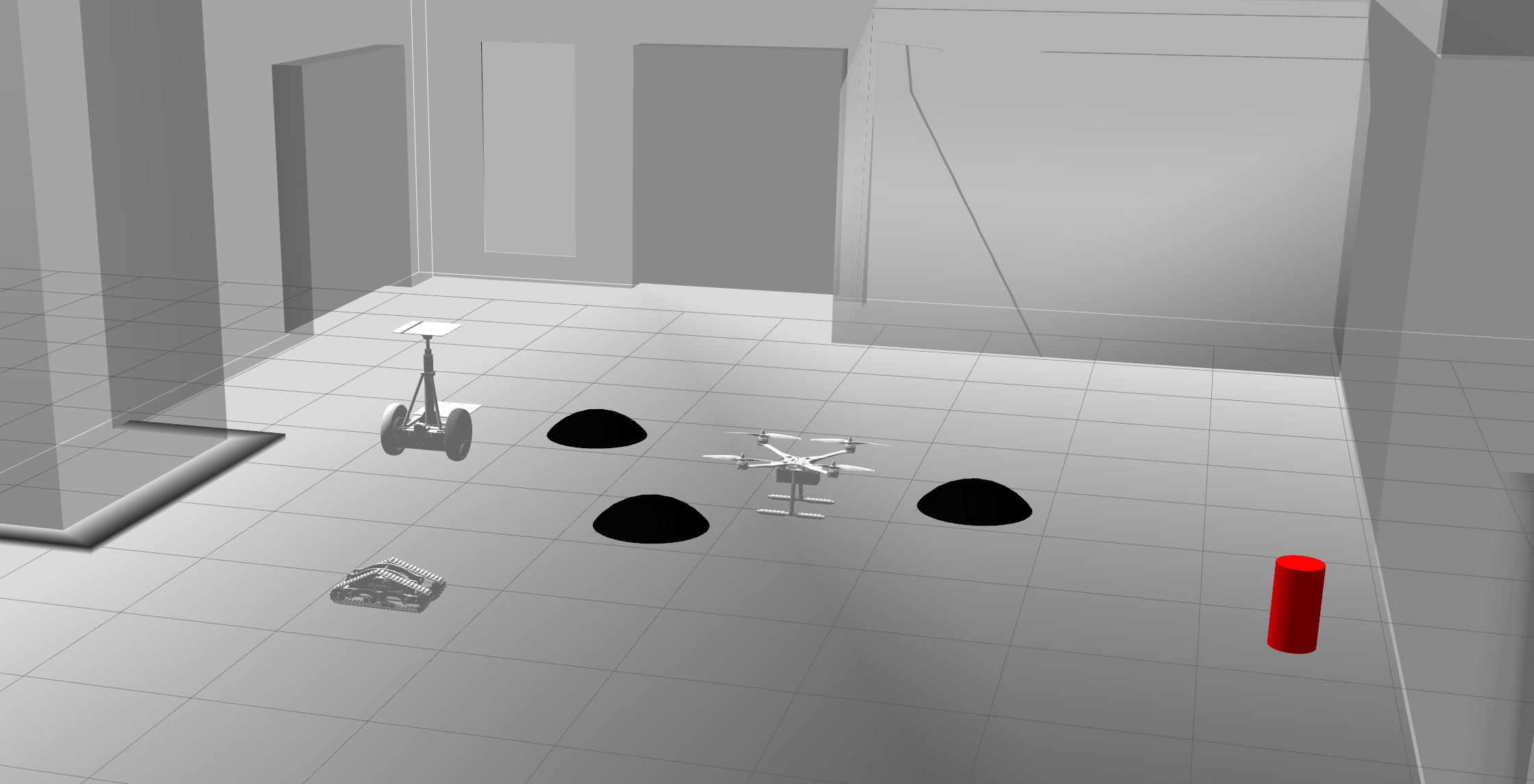}
    \caption{{The agents, the obtacles (black), and the sample (red)  in ROS simulation environment.}}
    \label{sim_overview}
\end{figure}

The set of agents includes the UAV, $A_U$, the Flipper, $A_F$, and the Segway, $A_S$. These agents all inhabit a planar $n \times m$ grid, with the drone located two meters above the ground vehicles. The beliefs of the vehicle locations in this grid are updated after each action is completed, based on the previous belief and the observation made. The initial vehicle locations are known, and the remaining system states are given by the environmental model. 

In order to capture the heterogeneity of the team, each grid in the environment has two states that measure the habitability of the grid for the Segway, as well as the probability that the grid in question contains the sample. These states are initialized to $0.5$, and observations of these states can be made when the Flipper or the UAV are within a certain distance from the cell. The Flipper can make accurate observations about the traversability of the terrain, but cannot sense the location of the sample well. The drone is more suited to locating the sample, but less suited to gauging traversability. 

The set of actions for each agent $A_i$ is the same, and consists of five actions: remaining in the same grid, and moving either forwards, backwards, left, or right to an adjacent grid. Thus, the total number of actions is $125$. The transitions between states are handled by controllers on the low-level dynamics, and the transition probability $T$ when moving from one grid to another is modeled as a high chance to move to the desired grid, and equal, smaller chances of landing in one of the eight grids adjacent to the desired grid. 

The components of the reward function for the Flipper and the UAV are measures of how much information will be gained from moving in that direction. The reward function also includes a heavy reward for the Segway moving towards a cell likely to contain the sample, and a heavy cost towards moving to a potentially dangerous cell. The observations for each agent update the environment states based on the observation made (binary detection) and the beliefs of the vehicles locations.

The exploratory mission is concluded when the sample has been collected, resulting in a mission success. {In terms of the system states, mission objective is satisfied when the Segway inhabits the same grid as the sample.} If the Segway enters an uninhabitable region, this results in a mission failure. Thus, the safe set of beliefs is defined as all states in which the Segway does not coincide with an uninhabitable region. For this mission, given the partial observability constraint, we require that there is a $95 \%$ probability of the Segway entering a habitable grid with each action. It is important to note that safety for this problem does not depend on the entirety of the belief space. Thus, it is possible to verify safety without computing the beliefs of each of the states.

\begin{figure*}
    \centering
    \includegraphics[width=5.5cm]{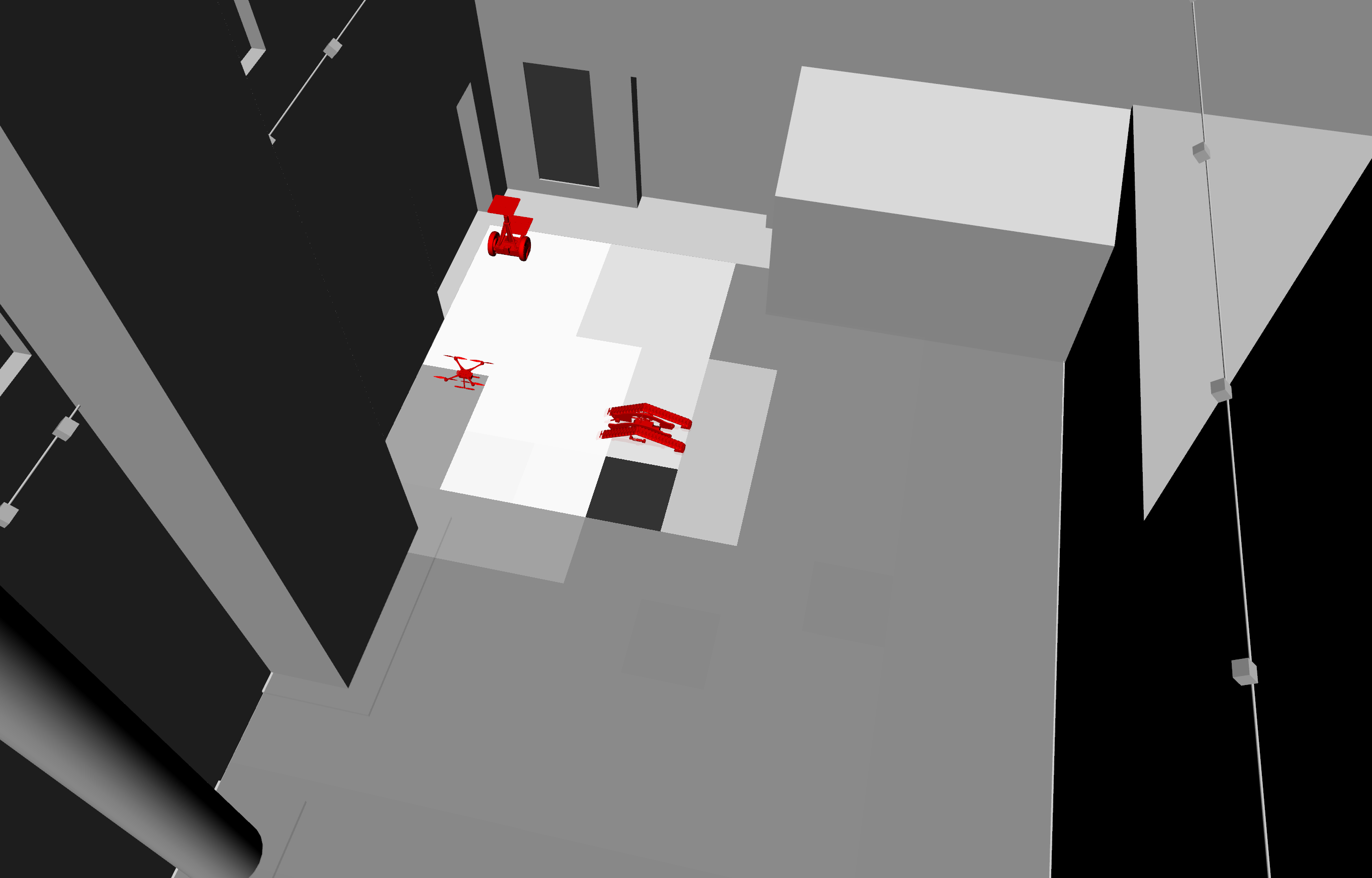}    \includegraphics[width=5.5cm]{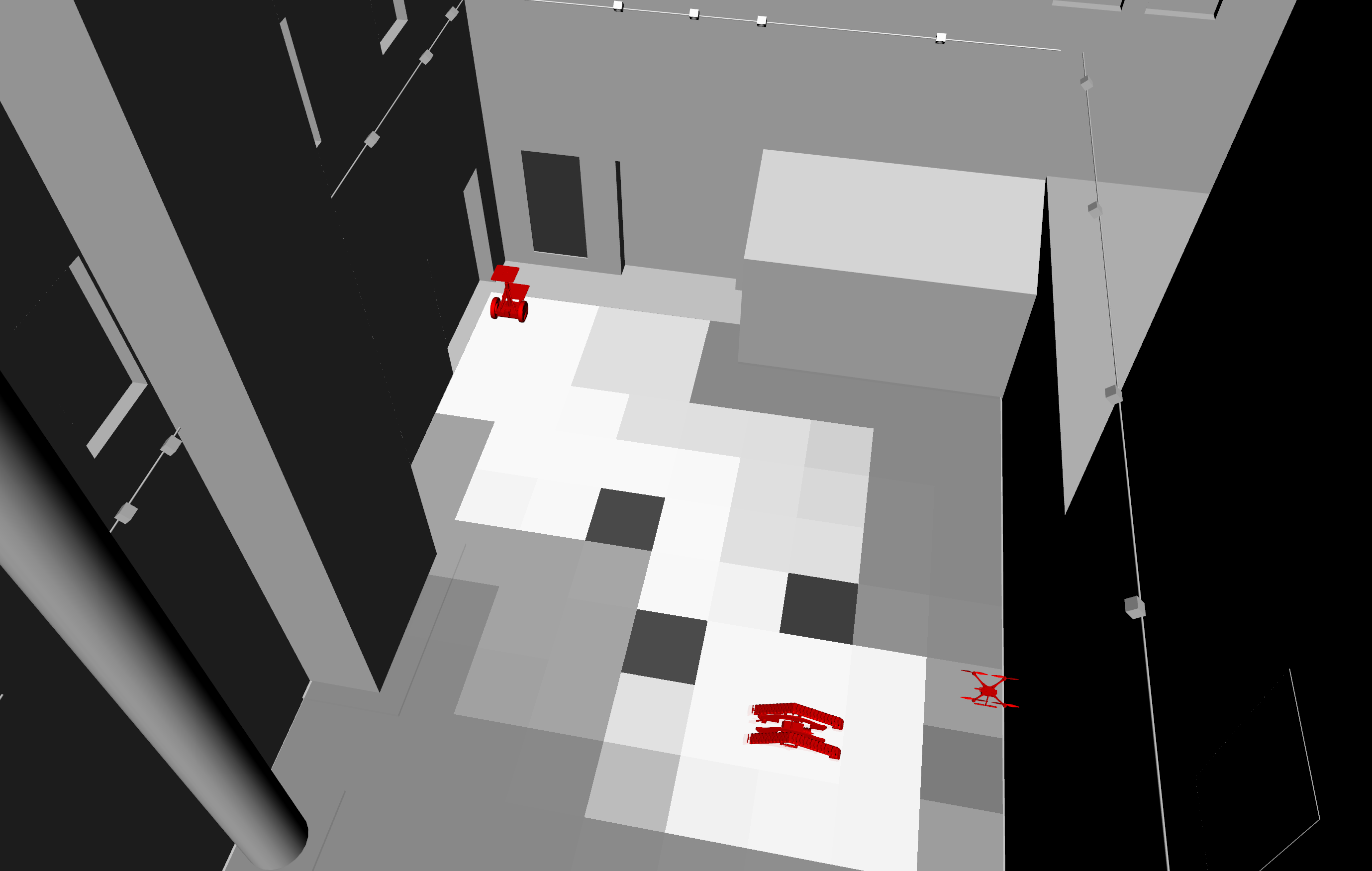}  \includegraphics[width=5.5cm]{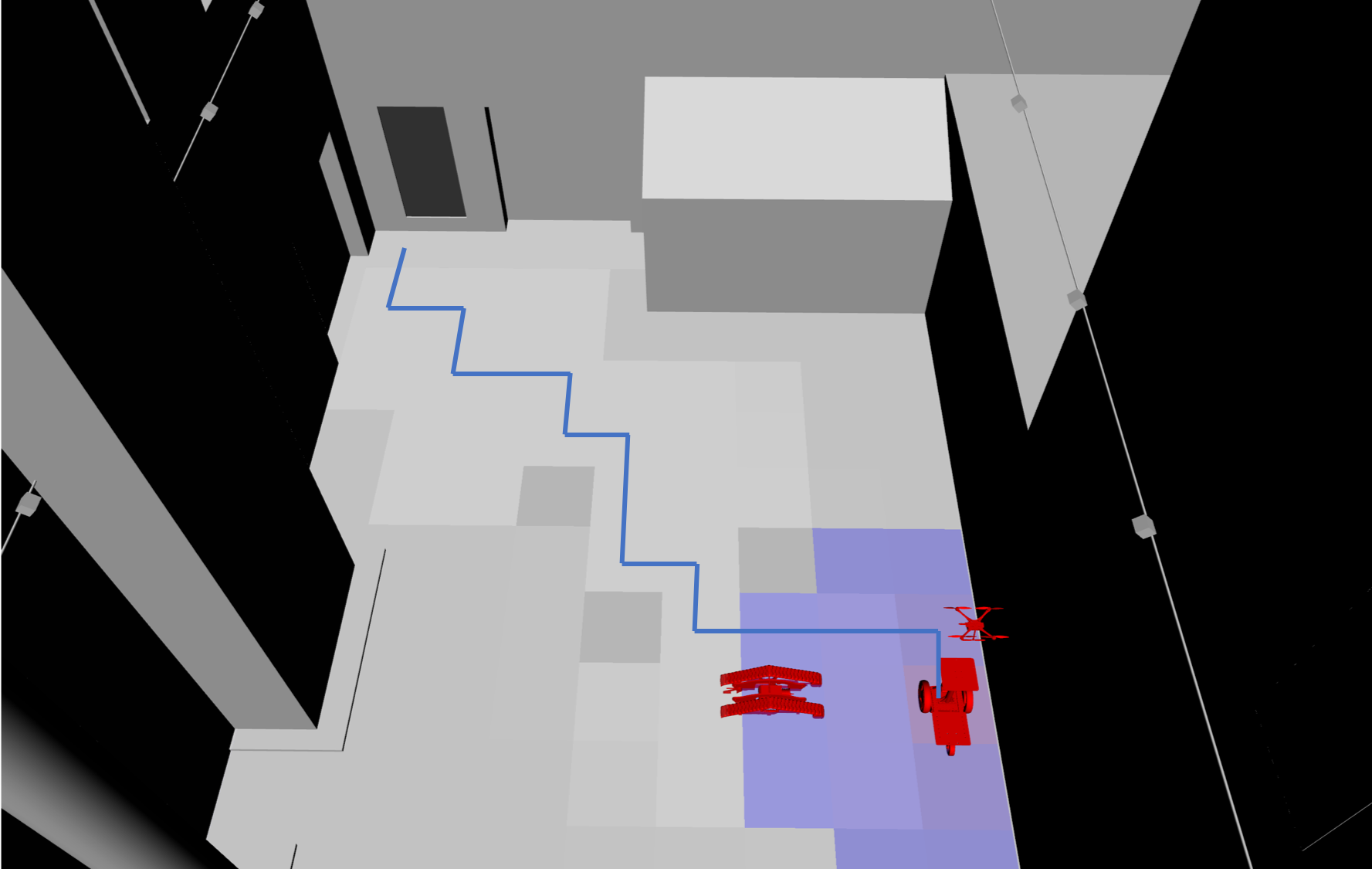}
    \caption{{Implementation of the nominal policy. Darker cells represent unsafe terrain,  and the blue cells in the third image represents the belief of the Segway location.}}
    \label{fig:nominal}
\end{figure*}

\begin{figure*}
    \centering
    \includegraphics[width=5.5cm]{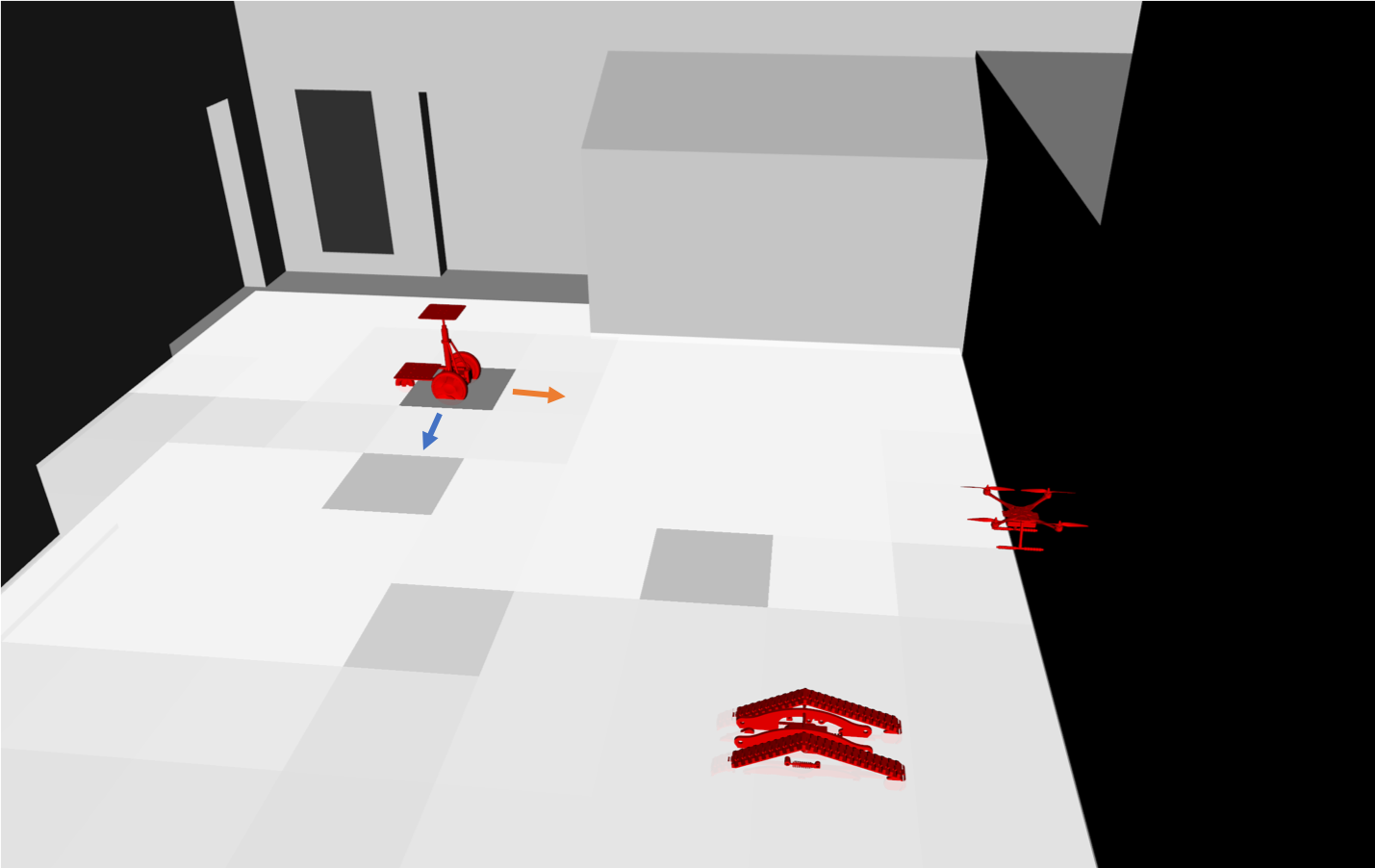}    \includegraphics[width=5.5cm]{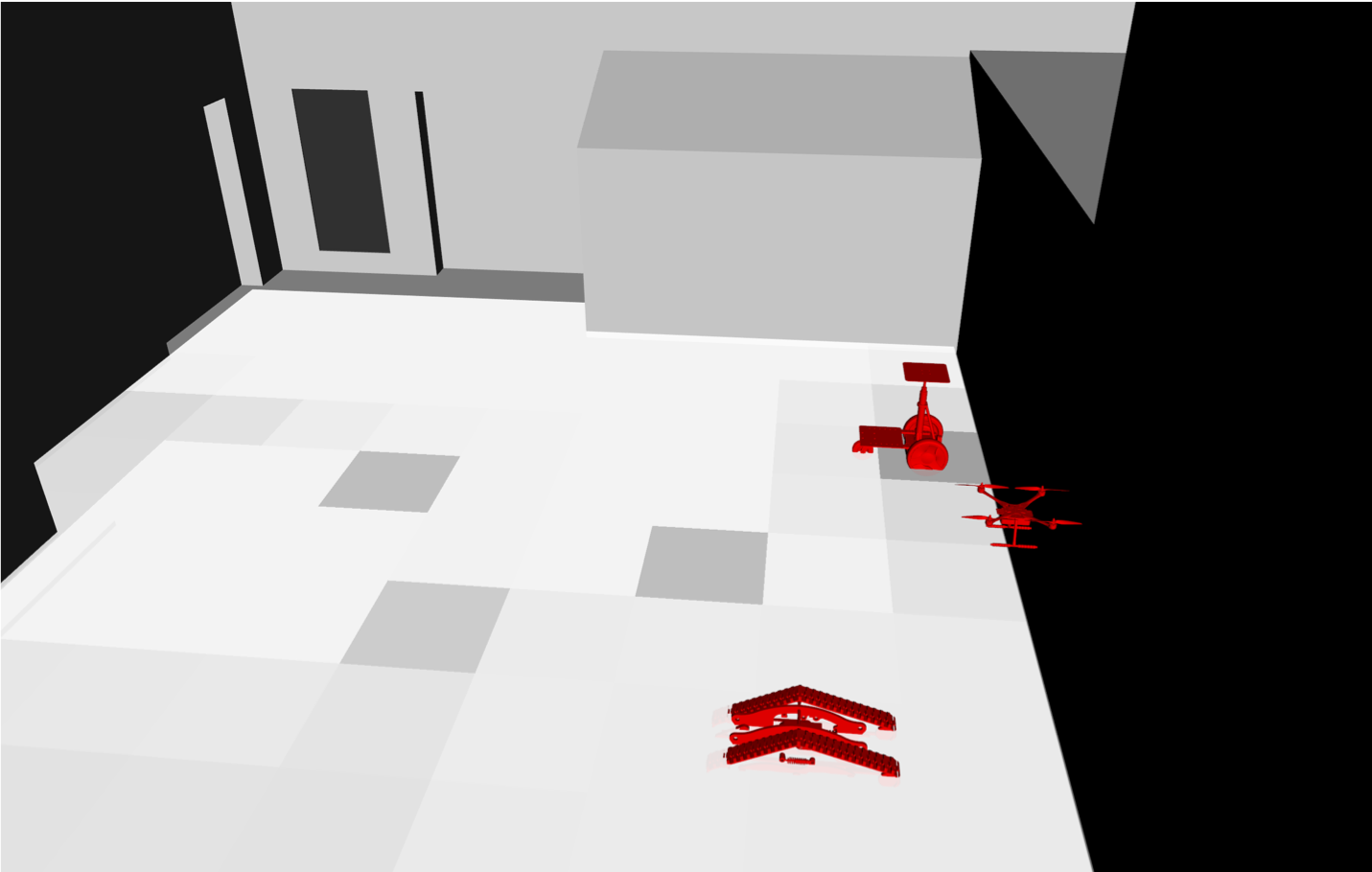}  \includegraphics[width=5.5cm]{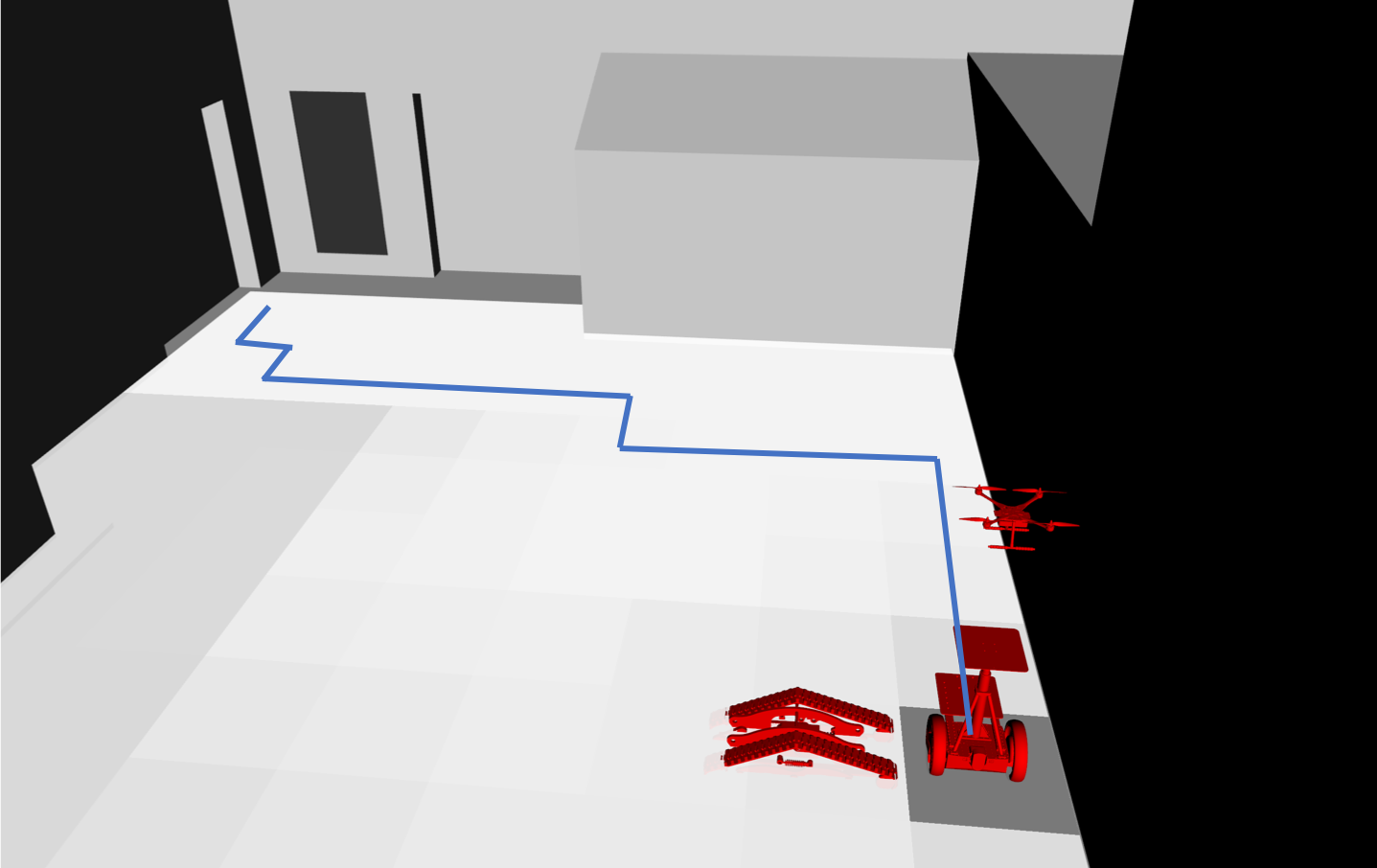}
    \caption{{Implementation of the safety filter. The blue arrow in the first image represents the desired action, and the orange arrow is the filtered action.}}
    \label{fig:filtered}
\end{figure*}

The simulation is carried out in a ROS environment as depicted in Figure~\ref{sim_overview}. Occupancy grids are utilized to represent the states of the system, which are updated after each action is taken. When an action is initialized, the low-level dynamics of the vehicles are simulated, and a message is published when the action is complete. Utilizing the observations made by this action, the beliefs are updated, and a new joint action is generated. The simulation is ended when the true position of the robot inhabits the same grid as the true position of the sample, or when the robot coincides with an uninhabitable cell.

To demonstrate the efficacy of the policy filter, a near-optimal policy that violates the belief safety filter was passed through Algorithm 3. The trajectory of the Segway under this policy is shown in Figure~\ref{fig:nominal}. While the policy is successful in simulation, due to perfect control over the states, it does not meet the imposed requirements for probabilistic $\%95$ safety. 

The resulting trajectory of the Segway after the policy filter is shown in Figure~\ref{fig:filtered}. The first filtered action occured at the time of the first image. While the desired trajectory of the Segway was to move towards the uninhabitable terrain, as shown in blue, the safety filter rejected this action. Instead, the action with the next highest reward, indicated by the orange arrow, was taken. This process continues, and the final trajectory is shown to move to the right wall of the building and then to the sample location. Thus, this filter was able to circumvent the unsafe policy, while still achieving the objective of the mission. While the resulting route is less optimal, it is a policy that could be implemented on a real system with realistic safety guarantees.


\section{Conclusions and Future Work}\label{sec:conclude}

We proposed a method for safe planning under uncertainty and partial observation of teams of heterogeneous robots modelled by MPOMDPs based on barrier functions. We applied our method for safe planning of a team of three robots using high-fidelity simulations. 

We considered agents with perfect communication. Prospective work will consider MPOMDPs with communication delays~\cite{oliehoek2012tree} or with no communication (decentralized POMDPs)~\cite{oliehoek2016concise}.

For Markov decision processes, safety can be encoded as a set of probabilistic temporal logic (PTL) specifications. In particular, in reinforcement learning finding optimal policies can induce unsafe behavior and shielded decision making~\cite{jansen2018shielded} has been introduced as an online safety filter to ensure safety in terms of PTL specifications. Future research will seek to present shielded decision making techniques for systems subject to uncertainty and partial observation using the DTBF-based safety filter developed in this paper.

Finally, in addition to high-fidelity simulations, we are  implementing the results discussed in this paper in the Center for Autonomous Systems  and Technologies (CAST) at the California Institute of Technology. Our eventual goal is to implement this work in the multi-agent planning framework for the DARPA Subterranean Challenge.  Our experimental observations will be disseminated in a follow up paper.






\bibliography{references}
\bibliographystyle{plain}

\end{document}